\documentclass[preprint]{article}

\usepackage[nonatbib]{neurips_2026}

\usepackage[utf8]{inputenc} 
\usepackage[T1]{fontenc}    
\usepackage{hyperref}       
\usepackage{url}            
\usepackage{booktabs}       
\usepackage{amsfonts}       
\usepackage{nicefrac}       
\usepackage{microtype}      
\usepackage{xcolor}         
\usepackage[numbers,sort&compress]{natbib}

\usepackage{amsmath}     
\usepackage{amssymb}     
\usepackage{amsfonts}    
\usepackage{mathtools}   
\usepackage{enumitem}

\title{Interpretability-Guided Layer Selection over Subspace Projection: SAEs as Stethoscopes, Not Scalpels, for Raw Task Vector Model Editing}

\author{%
  Li Lei \\
  Incept Labs\\
  Houston, TX\\
  \And
  Madalina Ciobanu \\
  Incept Labs\\
  Houston, TX\\
  \AND
  Qingqing Mao$^*$ \\
  Incept Labs, Houston, TX\\
  Titan Holdings, San Francisco, CA\\
  \texttt{qmao@inceptlabs.ai} \\
  \And
  Ritankar Das \\
  Incept Labs, Houston, TX\\
  Titan Holdings, San Francisco, CA\\
}

\begin{document}

\maketitle

\begin{abstract}
Large language models (LLMs) increasingly require surgical model
editing to enhance domain-specific capabilities without incurring
the computational cost or catastrophic forgetting associated with
full fine-tuning. Sparse Autoencoders (SAEs) have emerged as a
promising tool in this setting, in principle allowing for
feature-level identification of where to intervene. In this work, we rigorously evaluate an SAE-guided editing
pipeline for mathematical reasoning on Gemma-3-4B-IT and uncover
a fundamental failure mode:
\textbf{the intuitively appealing approach of projecting task
vectors onto SAE feature subspaces acts as an information
bottleneck that discards approximately 97\% of the modification
energy,} yielding no statistically significant improvements across
seven math subjects. We show that this failure stems from
a \emph{geometric misalignment} between activation-space SAE
directions and weight-space task vectors. We then propose a shift in
perspective: \textbf{SAE as a Stethoscope, Not a Scalpel}, where
SAEs are used for \emph{layer-level diagnosis} rather than
\emph{intervention-level filtering}. By injecting unfiltered raw
task vectors only into layers identified by an SAE-derived
specificity score, we improve Number Theory accuracy from 29.6\%
to 39.4\% ($z=+3.41$, $p=0.0007$) on the Minerva Math benchmark,
with 5 of 7 subjects significantly improved and none significantly degraded. Our method is fully deterministic, requires
no additional inference cost, and provides a principled framework
for interpretability-guided model editing.
\end{abstract}

\section{Introduction}
\label{sec:intro}

Large language models (LLMs) increasingly require \emph{surgical
model editing}: modifying only the components responsible for a
target capability while leaving everything else intact. Standard
adaptation methods such as LoRA~\citep{hu2022lora} modify weights
indiscriminately across all layers, leading to interference and
catastrophic forgetting. Two largely separate research threads
have pursued the surgical alternative.
\emph{Task arithmetic}~\citep{ilharco2023editing} treats the
weight difference between a fine-tuned model and its base as an
additive ``task vector'' that can be composed, negated, or
selectively reapplied in weight space.
\emph{Mechanistic interpretability} (MI) instead works in
activation space, decomposing hidden representations into
interpretable features---most prominently via Sparse Autoencoders
(SAEs)~\citep{bricken2023monosemanticity,cunningham2024sparse}---and
intervening on those features to steer model behavior.

\begin{figure}[t]
\centering
\includegraphics[width=\linewidth]{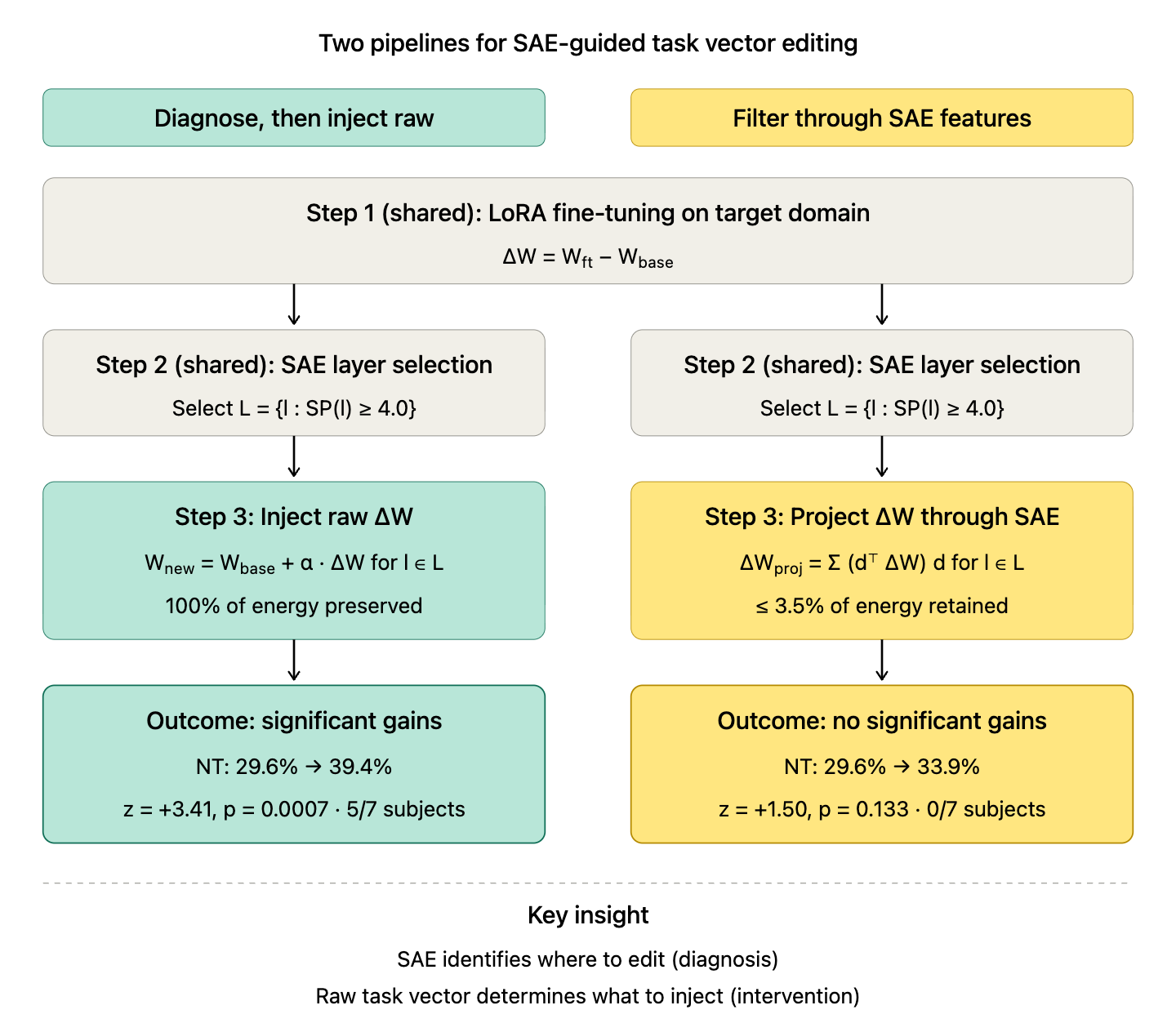}
\caption{Two pipelines for SAE-guided task vector model editing.
Both share Steps 1 (LoRA fine-tuning) and 2 (SAE layer
selection); they differ only in Step 3. \emph{Diagnose, then
inject raw} (left) injects the unfiltered task vector $\Delta W$
into SAE-selected layers, preserving 100\% of the modification
energy. \emph{Filter through SAE features} (right) projects
$\Delta W$ onto the subspace spanned by domain-specific SAE
decoder vectors, retaining only a few percent ($\leq 3.5\%$) of the energy. The
former produces statistically significant gains on 5 of 7 math
subjects; the latter produces none.}
\label{fig:pipelines}
\end{figure}

The intersection of these threads suggests a compelling pipeline:
use SAEs to identify domain-relevant features, then inject task
vectors \emph{only through those features}. Pre-trained SAEs at
scale, such as Gemma Scope~\citep{lieberum2024gemma} and Gemma
Scope 2~\citep{deepmind2025gemmascope2}, make this practical to
implement off the shelf. Figure~\ref{fig:pipelines} contrasts
this projection-based recipe with the alternative we propose.

\paragraph{A counterintuitive finding.}
We implement and rigorously evaluate this projection-based pipeline on
Gemma-3-4B-IT~\citep{gemma3team2025} for mathematical reasoning
and report a surprising negative result. \textbf{Projecting task
vectors onto SAE decoder subspaces retains only a few percent ($\leq \mathbf{3.5}\%$) of the energy (discarding the remainder to $\mathbf{100}\%$) and yields no statistically significant gains
on any of seven math subjects.} Increasing the SAE feature width
from 16K to 262K does not close the gap, indicating the
bottleneck is intrinsic to projection rather than feature
granularity. The failure has a clean mechanistic explanation:
SAE decoder vectors $d_{l,j}$ live in \emph{activation space} and
capture statistical regularities of how the model represents
information; task vectors $\Delta W$ live in \emph{weight space}
and encode how computations change. Projecting weight-space
modifications through activation-space directions conflates two
fundamentally different geometric structures, leading to
near-total loss of functional signal~\citep{sharkey2025open}.

This pattern is consistent with recent work questioning whether
SAE-based interventions are reliable beyond probing.
\citet{sharkey2025open} catalogue substantial limitations of
sparse dictionary learning. \citet{kantamneni2025sae} show that
SAE-based probes do not consistently outperform simple linear
baselines. \citet{heap2025random} find that SAEs trained on
randomly initialized transformers produce features
indistinguishable from those trained on real models. Concurrent
work on protein language models reports that activation-space
SAE editing and weight-space task arithmetic are
\emph{complementary} rather than substitutable, controlling
different properties~\citep{katki2025whereedit}. Together these
results suggest the activation/weight distinction is a
load-bearing axis of the editing problem.

\paragraph{SAE as stethoscope, not scalpel.}
Our central methodological claim is that SAEs are valuable as a \emph{diagnostic} tool, not an \emph{intervention} tool. We compute a per-layer specificity score ($SP$) derived from SAE feature analysis to identify domain-specialized layers, then selectively inject the unfiltered task vector into this subset. Notably, we identify an \textbf{empirical conservation of the modification budget}: as the number of selected layers decreases, the optimal scaling parameter $\alpha$ increases proportionally to maintain functional impact ($n_{layers} \times \alpha_{opt} \approx 11.2$). This separation preserves 100\% of the modification energy in target layers while confining edits to the model's domain-relevant regions.

\paragraph{Contributions.}
\textbf{(i)} We propose \emph{SAE-Guided Layer Selection with Raw
Task Vector Injection} and show on Minerva
Math~\citep{hendrycks2021math} that it improves Number Theory
accuracy from $29.6\%$ to $39.4\%$ ($z = +3.41$, $p = 0.0007$),
with 5 of 7 math subjects significantly improved and none significantly
degraded. \textbf{(ii)}  We document a substantive negative
result: SAE projection of task vectors discards $\sim$97\% of the
modification energy and produces no significant gains, even with
a $16\times$ larger SAE---consistent with recent concerns about
SAEs as intervention
tools~\citep{sharkey2025open,kantamneni2025sae,heap2025random}.
This negative result is the central scientific contribution of
this work; the positive method
(Section~\ref{sec:method}) is a constructive corollary that
demonstrates SAE diagnostic utility separately from the failed
projection approach.
\textbf{(iii)} We provide $30+$ ablation configurations,
including alpha response curves, layer-selection strategies,
multiple task vector sources, and dual-domain injection;
extended ablations and reproduction details are in the appendix.
We discuss limitations in Section~\ref{sec:limitations}.

\section{Related Work}
\label{sec:related}

\paragraph{Task arithmetic and weight-space model editing.}
\citet{ilharco2023editing} introduced task vectors as the weight difference between models. Subsequent work addresses interference when combining multiple vectors, including TIES-Merging~\citep{yadav2023ties}, DARE~\citep{yu2024language}, Localize-and-Stitch~\citep{he2025localize}, and Subspace Boosting~\citep{skorobogat2025subspace}. While ROME~\citep{meng2022locating} and MEMIT~\citep{meng2023memit} utilize \emph{causal tracing} to locate factual associations, our method provides a \textbf{fully deterministic, training-free diagnostic} for identifying domain-level specialization. This aligns with the findings of \citet{ortizjimenez2023tangent}, who established that \textbf{weight disentanglement} is the crucial factor for effective task arithmetic. Our SAE-guided selection provides a principled criterion for isolating these task-relevant updates.

\paragraph{Sparse autoencoders and their limitations.}
SAEs decompose neural network activations into sparse,
monosemantic features~\citep{bricken2023monosemanticity,cunningham2024sparse},
with comprehensive open suites available for the Gemma
family~\citep{lieberum2024gemma, deepmind2025gemmascope2} and Llama
family~\citep{he2024llamascope}. Recent work documents
fundamental limitations: SAE reconstruction systematically loses
information~\citep{sharkey2025open}; concepts may be convex
regions rather than linear directions~\citep{fel2026rabbit};
SAEs assume i.i.d.\ activations, ignoring sequential
dependencies relevant to chain-of-thought
reasoning~\citep{lubana2025priors}; SAE-based probes do not
consistently outperform simple
baselines~\citep{kantamneni2025sae}; and SAEs trained on
randomly initialized transformers learn features qualitatively
similar to those from trained models~\citep{heap2025random}.

\paragraph{Weight-space vs.\ activation-space interventions.}
JoLA~\citep{lai2025jola} jointly learns intervention locations
and parameters in activation space.
\citet{katki2025whereedit} find that for protein language models,
weight-space task arithmetic and activation-space SAE editing are
complementary, each more effective for a different subset of
biochemical properties. Our paper contributes a sharp empirical
demonstration of when the two spaces are \emph{not}
interchangeable: a weight-space modification projected through an
activation-space basis loses approximately 97\% of its energy.

\section{Method}
\label{sec:method}

\paragraph{Notation.}
Let $W_{\text{base}}, W_{\text{ft}} \in \mathbb{R}^d$ denote the
parameters of the base and fine-tuned models. Following
\citet{ilharco2023editing}, the \emph{task vector} is the
element-wise weight difference
$\Delta W = W_{\text{ft}} - W_{\text{base}}$. We write $W^{(l)}$
for the slice corresponding to layer
$l \in \{1,\dots,L\}$. A Sparse Autoencoder at layer $l$ provides
an encoder--decoder pair $(E_l, D_l)$ with $D$ latent features;
each column $d_{l,j}$ of $D_l$ is interpreted as a feature
direction. We use Gemma Scope 2~\citep{deepmind2025gemmascope2} with
$D = 16{,}384$ features per layer in our primary experiments and
$D = 262{,}144$ in our scale ablation.

Our method (Figure~\ref{fig:pipelines}) consists of three steps:
(i) extract a task vector via domain-specific LoRA fine-tuning;
(ii) use SAE feature analysis to identify domain-specialized
layers; and (iii) inject the raw, unfiltered task vector only
into those layers.

\paragraph{Task vector extraction.}
We obtain $\Delta W$ via LoRA fine-tuning~\citep{hu2022lora} on a
domain-specific dataset. The selective injection pipeline is
agnostic to the source of $\Delta W$ and would apply equally to
task vectors from full fine-tuning.

\begin{figure}[t]
\centering
\includegraphics[width=1.0\linewidth]{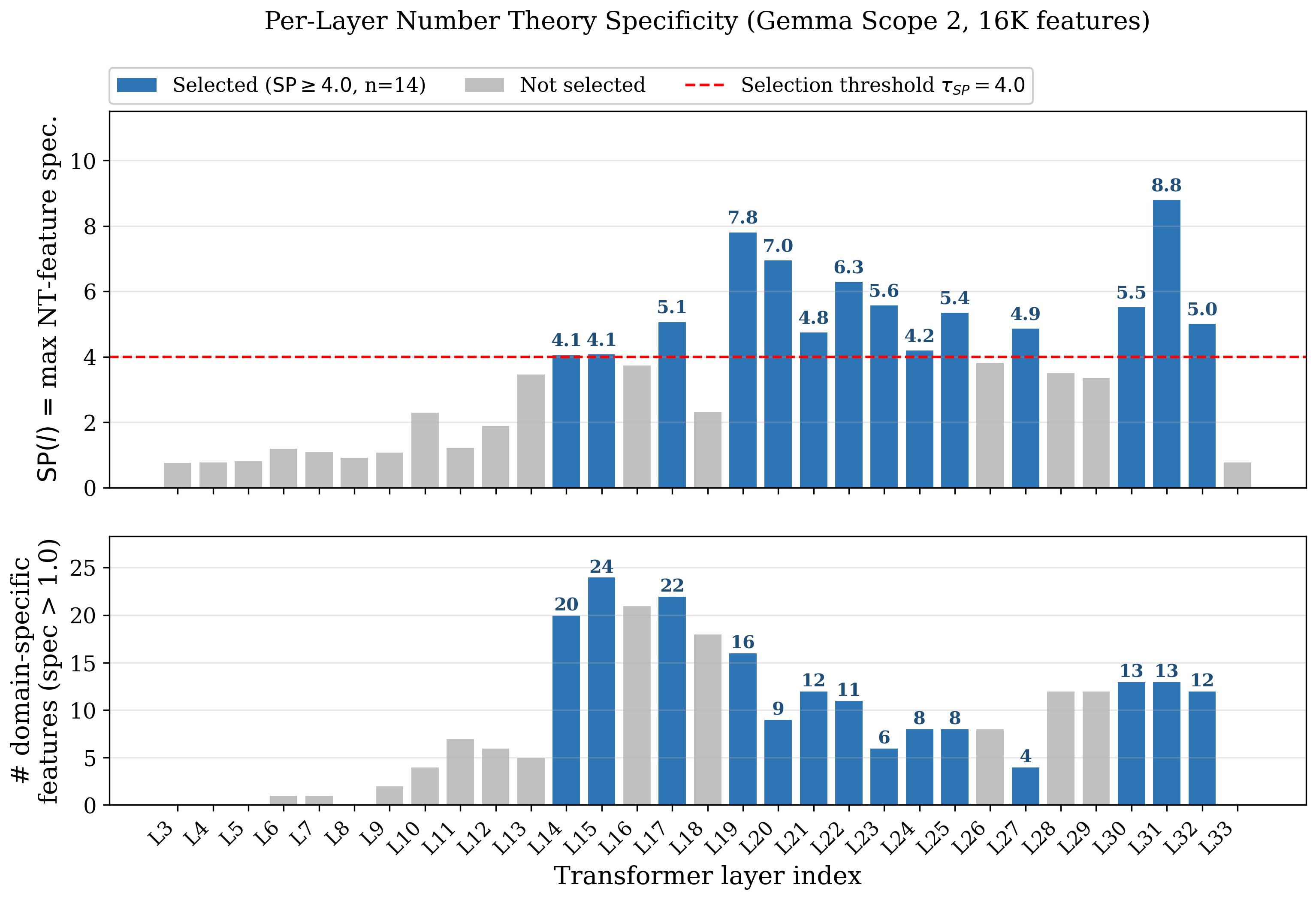}
\caption{Per-layer Number Theory specificity (Gemma Scope 2, 16K features).
\textit{Top}: Specificity Score $\mathrm{SP}(l) = \max_j \mathrm{spec}(l,j)$
per layer; blue bars mark the 14 layers selected at
$\tau_{\mathrm{SP}} = 4.0$, the dashed red line indicates the threshold.
\textit{Bottom}: Number of domain-specific features
($\mathrm{spec}(l,j) > 1.0$) per layer. The distribution is bimodal:
a mid-network cluster (layers 14--17) and a late-network cluster
(layers 30--32) carry the strongest domain specialization.}
\label{fig:sp_scores}
\end{figure}

\paragraph{SAE-guided layer selection.}
For each layer $l$ and SAE feature $j$, we compare the feature's
mean activation on target-domain inputs to its mean activation
on all other domains:
\begin{equation}
\mathrm{spec}(l, j) = \frac{\bar{a}^{\text{target}}_{l,j}}
{\bar{a}^{\text{other}}_{l,j} + \epsilon},
\label{eq:feature-spec}
\end{equation}
with $\epsilon = 10^{-6}$. A feature is \emph{domain-specific} if
$\mathrm{spec}(l, j) > \tau_f$ (we use $\tau_f = 1.0$). We aggregate feature-level specificity into a per-layer Specificity Score by taking the \textit{maximum} rather than a count. Empirically the maximum is more discriminative than the count: it captures whether \textit{any} highly-specialized feature exists in a layer, not merely how many low-specificity features happen to exceed a soft threshold. A layer with a single feature at spec=10 is more domain-relevant than a layer with twenty features at spec=1.5. The threshold $\tau_{\mathrm{SP}}=4.0$ was chosen on a held-out validation split and is ablated in Section~\ref{sec:results-layers}.
\begin{equation}
  \mathrm{SP}(l) = \max_{j \in \{1,\dots,D\}} \mathrm{spec}(l, j),
  \label{eq:sp_score}
\end{equation}
the maximum specificity over all SAE features in layer $l$.
We select layers with $\mathrm{SP}(l) \geq \tau_{\mathrm{SP}}$,
using $\tau_{\mathrm{SP}} = 4.0$ in our main experiments,
which yields 14 of 34 layers (Figure~\ref{fig:sp_scores}).

\paragraph{Selective raw task vector injection.}
Given selected layers $\mathcal{L}_{\text{selected}}$ and a
global scaling parameter $\alpha$:
\begin{equation}
W^{(l)}_{\text{new}} =
\begin{cases}
W^{(l)}_{\text{base}} + \alpha \cdot \Delta W^{(l)}
& \text{if } l \in \mathcal{L}_{\text{selected}} \\
W^{(l)}_{\text{base}} & \text{otherwise}.
\end{cases}
\label{eq:inject}
\end{equation}
$\Delta W^{(l)}$ is the \emph{unfiltered} per-layer task vector:
no projection, no truncation, no SAE-based masking. Once
applied, the edited model has identical architecture, parameter
count, and inference cost to the base model. The optimal
$\alpha$ is determined via grid search.

\paragraph{Why not project task vectors through SAE features?}
A natural alternative projects $\Delta W$ onto the subspace
spanned by domain-specific SAE decoder vectors,
$\Delta W^{(l)}_{\text{proj}} = \sum_{j \in \mathcal{F}_l}
(d_{l,j}^\top \Delta W^{(l)} d_{l,j}) /
(d_{l,j}^\top d_{l,j})$,
where $\mathcal{F}_l = \{j : \mathrm{spec}(l, j) > \tau_f\}$. Recall that task vectors $\Delta W$ modify the weights directly, while SAE decoder features $d_{l,j}$ describe directions in activation space, so any projection of $\Delta W$ through $d_{l,j}$ necessarily mixes two distinct spaces. In practice this retains only a few percent (roughly $2$--$3.5\%$ across our SAE configurations) of the modification
energy ($\|\Delta W^{(l)}_{\text{proj}}\|_F /
\|\Delta W^{(l)}\|_F$) and systematically degrades performance
(Section~\ref{sec:results}). The failure mode is geometric: SAE
decoder vectors span the activation manifold at layer $l$,
which is a different geometric object than the column space of
$\Delta W^{(l)}$. Substituting one for the other introduces a
representational mismatch~\citep{sharkey2025open}. We note that the near-total energy loss reported in Section~\ref{sec:results} is the primary empirical evidence for this geometric mismatch; the mechanistic claim and the experimental observation are mutually consistent rather than independently derived.

\section{Experimental Setup}
\label{sec:setup}

\paragraph{Model and SAE.}
Gemma-3-4B-IT~\citep{gemma3team2025}, a 4B-parameter
instruction-tuned transformer with 34 layers. Gemma Scope
2~\citep{deepmind2025gemmascope2} provides pre-trained SAEs at every
layer, with $16$K features in main experiments and $262$K in our
scale ablation.

\paragraph{Task vector.}
Our primary task vector (v2) is obtained by LoRA fine-tuning on
$3{,}865$ Number Theory problems from the MATH training
set~\citep{hendrycks2021math}, with rank $r = 16$, learning rate
$2 \times 10^{-4}$, and 5 epochs. Two alternatives (v25:
$6{,}155$ samples; v3: $20{,}000$ mixed-domain) test sensitivity
to the source of $\Delta W$ (see appendix). For comparison
against SAE projection in Section~\ref{sec:results-projection},
we additionally define a smaller raw-injection control. The E3 baseline applies the raw task vector to a fixed seven-layer subset
$\{19,20,22,23,25,30,31\}$ (configuration \texttt{E3\_7L} in our codebase), chosen as the seven highest-\textsc{SP} layers among NT-analyzed layers in our initial sweep, rather than an \textsc{SP} threshold applied independently of rank. E3 isolates the contribution of layer \emph{count} from the contribution of raw versus projected injection.

\paragraph{Evaluation.}
Minerva Math via lm-evaluation-harness~\citep{gao2024lmeval}: 540
problems across 7 subjects (NT, CP, ALG, GEO, IA, PRE, PC). We
use 4-shot prompting with greedy decoding and the
\texttt{math\_verify} metric.

\paragraph{Baselines.}
Three baselines share the same v2 task vector:
\textbf{(i) Base} (unmodified Gemma-3-4B-IT);
\textbf{(ii) Full LoRA merge} (add $\Delta W$ to all 34 layers);
\textbf{(iii) SAE Projection} (inject $\Delta W_{\text{proj}}$
into the same SP-selected layers). Each baseline's $\alpha$ is
independently grid-searched.

\paragraph{Statistical testing.}
Per subject we compute a two-sample proportion $z$-test:
$z = (\hat{p}_{\mathrm{edit}} - \hat{p}_{\mathrm{base}}) /
\sqrt{\mathrm{SE}_{\mathrm{edit}}^2 + \mathrm{SE}_{\mathrm{base}}^2}$,
with two-sided $p$-values and significance at $|z| \geq 1.96$.
Sample sizes are the full Minerva Math test sets:
$n=540$ (NT), $474$ (CP), $1{,}187$ (ALG), $479$ (GEO),
$903$ (IA), $871$ (PRE), $546$ (PC).
With these sample sizes the test is well-powered
to detect effects of approximately 4 percentage points on every subject.

\paragraph{Determinism and hardware.}
All evaluations use greedy decoding with deterministic weight
modifications; repeated runs of the same configuration produce
identical accuracies at float64 precision (verified;
Section~\ref{sec:analysis}). Experiments ran on $3 \times$ H100
80GB GPUs, totaling $\sim 290$ GPU-hours across reported
configurations, plus an additional $\sim 60$ GPU-hours for the
exploratory CMA-ES search reported in Appendix~\ref{app:ppl} (Figure~\ref{fig:ppl-scatter}).

\section{Results}
\label{sec:results}

We organize results in four parts. Section~\ref{sec:results-main}
presents the main comparison against three baselines on Minerva
Math (Table~\ref{tab:main}, Figure~\ref{fig:bar-comparison}).
Sections~\ref{sec:results-alpha}--\ref{sec:results-projection}
isolate three design choices: the scaling parameter $\alpha$,
layer selection strategy, and---most critically---raw versus
SAE-projected task vectors. Three findings stand out:
(i) selective injection into 14 SAE-identified layers improves 5
of 7 math subjects with no degradation; (ii) performance is
robust across a wide range of $\alpha$ but sensitive to layer
selection above the threshold $\tau_{\text{SP}} = 4$; and
(iii) projecting $\Delta W$ through SAE feature subspaces
produces no statistically significant gains on any subject,
despite retaining the same layer selection.

\subsection{Main Results}
\label{sec:results-main}

Table~\ref{tab:main} compares our method against three baselines
sharing the same task vector source (v2). Our method achieves
statistically significant improvements on 5 of 7 subjects, with
the primary target (Number Theory) showing the strongest effect
($z = +3.41$, $p = 0.0007$, $+9.81$ pp absolute, $+33\%$
relative). Crucially, \emph{no subject is degraded}: Intermediate Algebra is positive but does not reach significance ($p=0.099$ two-sided); PreCalculus is unchanged. This demonstrates that selective injection avoids
catastrophic forgetting without explicit regularization.

\begin{table}[t]
\centering
\caption{Performance on Minerva Math ($7$ subjects). Accuracy
(\%) and $z$-scores vs.\ base. \textbf{Bold}: statistically
significant ($z \geq 1.96$, $p < 0.05$). $^{\dagger}$: best
SAE-projected configuration (rank-3, $\alpha=0.70$, 16K SAE).
Ours: $\mathrm{SP}\geq 4.0$ layer selection (14/34 layers),
$\alpha = 0.80$.}
\label{tab:main}
\small
\begin{tabular}{lccccccccc}
\toprule
Method & NT & CP & ALG & GEO & IA & PRE & PC & \#Sig \\
\midrule
Base model            & 29.6 & 33.3 & 61.3 & 30.5 & 16.6 & 62.9 & 20.5 & --  \\
Full LoRA merge (34L) & 30.0 & 33.8 & 58.1 & 29.2 & 16.3 & 61.4 & 19.8 & 0/7 \\
SAE Projection$^{\dagger}$ & 33.9 & 34.6 & 62.1 & 31.0 & 16.8 & 63.5 & 20.3 & 0/7 \\
Ours (SP4, $\alpha{=}0.80$) & \textbf{39.4} & \textbf{41.4} & \textbf{67.0} & \textbf{37.2} & 19.6 & \textbf{69.6} & 20.5 & \textbf{5/7} \\
\midrule
$z$-score & $+3.41$ & $+2.56$ & $+2.87$ & $+2.19$ & $+1.65$ & $+2.94$ & $+0.00$ & \\
$p$-value & $0.0007$ & $0.0105$ & $0.0041$ & $0.0286$ & $0.0989$ & $0.0032$ & $1.0000$ & \\
\bottomrule
\end{tabular}
\end{table}

\begin{figure}[t]
\centering
\includegraphics[width=1.0\linewidth]{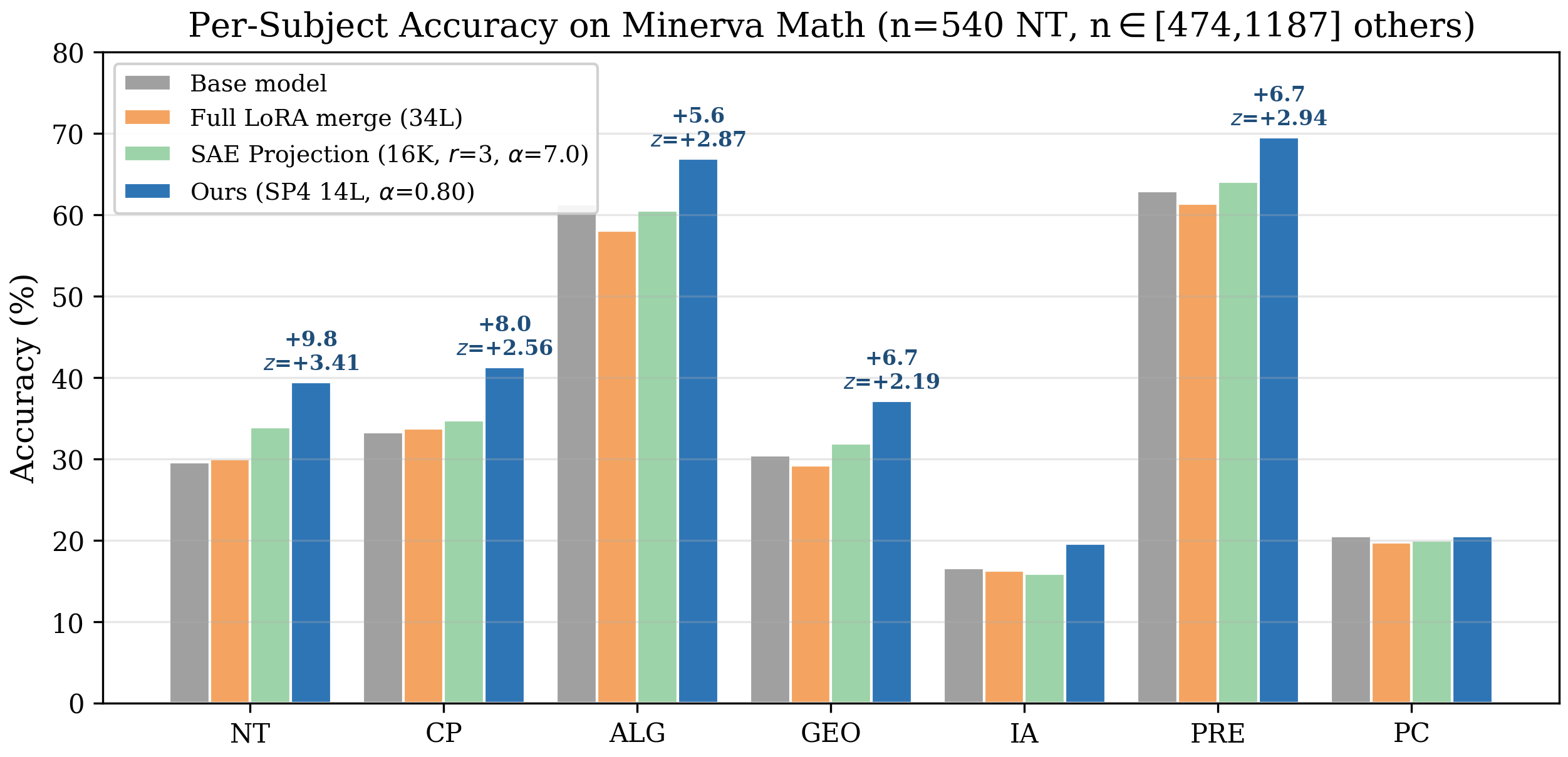}
\caption{Per-subject accuracy across the seven Minerva Math
subjects ($n=540$ NT, $n \in [474, 1{,}187]$ others).
Four conditions are shown left to right: Base model (leftmost/gray),
Full LoRA merge 34L (middle left/orange), SAE Projection 16K (middle right/green), and Ours SP4 14L $\alpha{=}0.80$ (rightmost/blue). Our method achieves statistically significant gains on 5 of 7 subjects, with absolute improvements ranging from $+5.6$ pp on Algebra to $+9.8$ pp on Number Theory. Full LoRA merge degrades three subjects; SAE Projection (green bars) produces numerical shifts that do not reach significance on any subject, despite using the same layer-selection signal.}
\label{fig:bar-comparison}
\end{figure}

Two baselines fail to produce significant improvement on any
subject. Full LoRA merge across all 34 layers actually
\emph{degrades} performance on Algebra ($-3.2$ pp), Geometry
($-1.3$ pp), and PreAlgebra ($-1.5$ pp), consistent with prior
findings that uniform task-vector application introduces
interference~\citep{yadav2023ties}. SAE Projection shows
numerical improvements but none reach significance, despite
using the \emph{same layer-selection signal} as our
method---confirming that the projection step itself, not layer
selection, is the bottleneck (Figure~\ref{fig:bar-comparison}).

\subsection{Robustness to the Scaling Parameter}
\label{sec:results-alpha}

We searched $\alpha \in \{0.70, 0.75, \dots, 1.20\}$ for the SP4
14L configuration. The response surface is broad and flat: NT
$z$-scores remain above $+2.84$ across the full range
$\alpha \in [0.70, 1.10]$ (a $40\%$ swing in scaling magnitude),
with the peak at $\alpha = 0.80$ ($z = +3.41$) only marginally
above the next-best values at $\alpha = 0.85$ ($z = +3.34$) and
$\alpha = 0.70, 1.10$ ($z = +3.16$). Performance drops at
$\alpha = 1.20$ ($z = +2.08$), marking the onset of
over-modification. The empirical relationship
$n_{\text{layers}} \times \alpha_{\text{opt}} \approx 11.2$
($14 \times 0.80$) for our configurations is consistent with a conserved ``modification budget'' across layer-selection strategies:
Figure~\ref{fig:alpha-curve} shows the full SP4~14L response
surface with the robust plateau annotated, and
Figure~\ref{fig:alpha-two-configs} shows that SP4~noDeep (11
layers) reaches its peak at $\alpha^* = 1.00$, preserving the
product $n_l \times \alpha^* \approx 11$. Full sweep and both
figures in Appendix~\ref{app:alpha}.

\subsection{Layer Selection Strategies}
\label{sec:results-layers}

We compare six layer-selection strategies, each at its respective
optimal $\alpha$ (Table~\ref{tab:layer-strategies}). Three
findings:

\paragraph{The SP threshold matters more than layer count.}
Three configurations select roughly the same number of layers
(11--14) but produce wildly different outcomes. SP4 14L
($z = +3.41$) and SP4 noDeep ($z = +3.22$) are both strong;
SP4.5 11L collapses to $z = +0.73$ despite a similar layer
count. Inspecting the SP4.5 selection reveals that raising the
threshold from $4.0$ to $4.5$ excludes layers $14$, $15$, and
$24$, which carry essential domain computation. Lowering the
threshold to SP $\geq 3.5$ (17 layers) dilutes the signal with
weakly specialized layers ($z = +2.33$). The SP threshold acts
as a categorical separator between domain-specialized and
non-specialized layers, not a smooth knob.

\paragraph{Deep layers carry domain-specific but interference-prone modifications.}
Removing layers 30--32 trades NT performance ($z = +3.22$ vs.\
$+3.41$) for broader coverage (6/7 vs.\ 5/7 significant
subjects), suggesting deep-layer task vectors encode NT-specific
computations that mildly conflict with Intermediate Algebra.

\paragraph{Mid-layer heuristics are insufficient.}
The MID 10L configuration (a hand-chosen middle slab, layers
17--27) achieves only $z = +2.59$ on NT and 3/7 significant
subjects. SAE-derived specificity identifies layers outside this
range (notably 14, 15, and 30--32) that are critical to the
effect. Per-subject $z$-scores across all six configurations are
shown in Figure~\ref{fig:layer-heatmap} (Appendix~\ref{app:layers}).

\begin{table}[t]
\centering
\caption{Layer-selection strategy comparison. SPx denotes layers
with Specificity Score $\geq x$. MID: layers $17$--$27$.
SP4\_noDeep: SP4 with deep layers $30$--$32$ removed. Each
configuration uses its optimal $\alpha$. Per-subject heatmap in
Appendix~\ref{app:layers}.}
\label{tab:layer-strategies}
\small
\begin{tabular}{lcccll}
\toprule
Configuration & \#Layers & $\alpha^{*}$ & NT $z$ & \#Sig & Note \\
\midrule
SP4 14L     & 14 & 0.80 & $+3.41$ & 5/7 & NT-best \\
SP4 noDeep  & 11 & 1.00 & $+3.22$ & 6/7 & Multi-subject best \\
SP3.5 17L   & 17 & 1.00 & $+2.33$ & 6/7 & Signal diluted \\
SP4 UNION   & 16 & 1.00 & $+2.33$ & 6/7 & Low-SP noise \\
MID 10L     & 11 & 0.90 & $+2.59$ & 3/7 & Missing deep layers \\
SP4.5 11L   & 11 & 1.00 & $+0.73$ & 0/7 & Excludes critical layers \\
\bottomrule
\end{tabular}
\end{table}

\subsection{Raw Task Vectors versus SAE Projection}
\label{sec:results-projection}

This section presents the central empirical claim of the paper:
SAE projection systematically underperforms raw injection of the
\emph{same} task vector through the \emph{same} layers
(Table~\ref{tab:projection}; cf.\ Figure~\ref{fig:pipelines} for
the conceptual contrast), and across all 30+ configurations we
tested no SAE-projected variant achieved a statistically significant
gain on Number Theory.

\begin{figure}[t]
\centering
\includegraphics[width=1.0\linewidth]{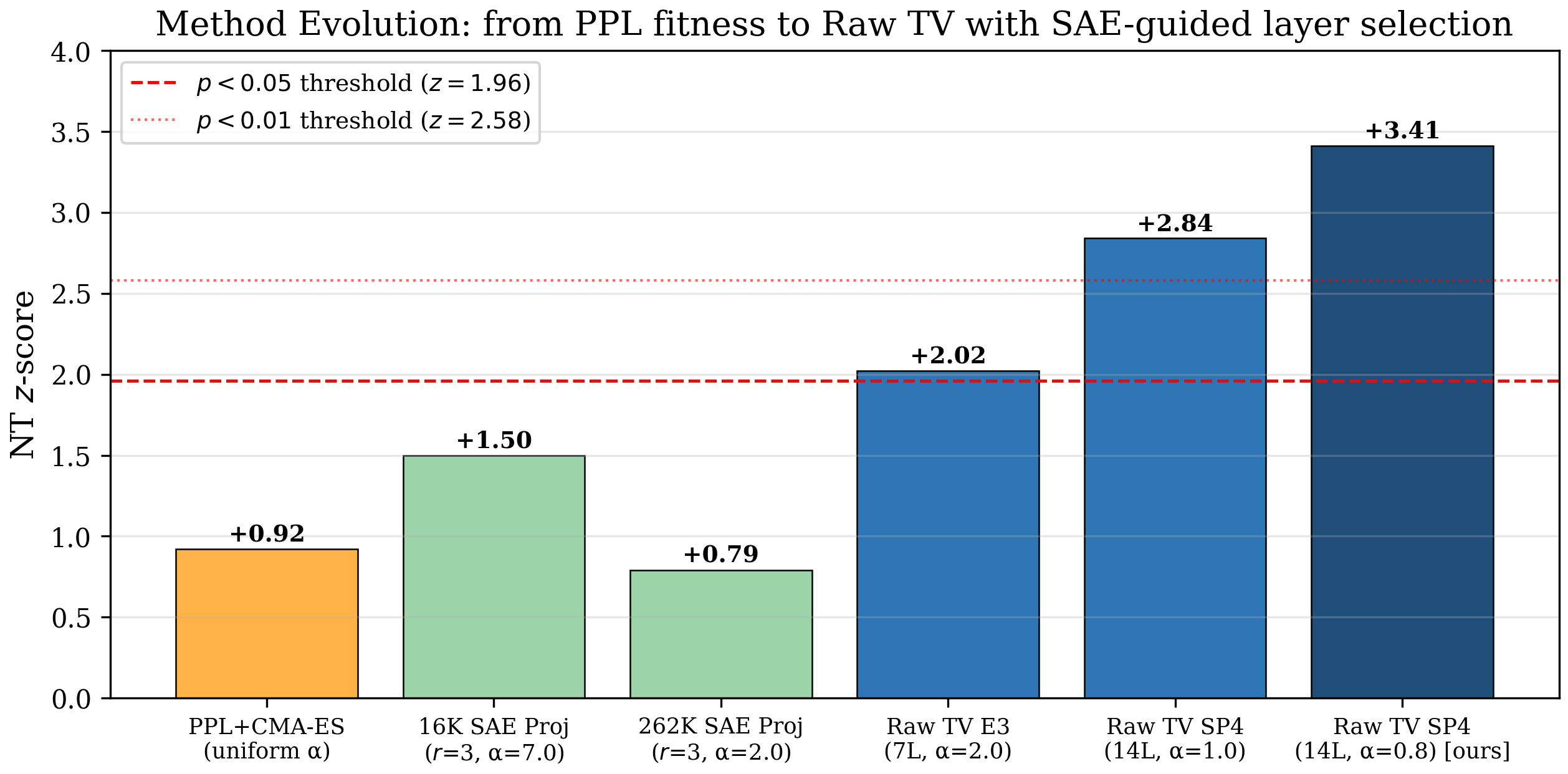}
\caption{Method evolution: NT $z$-score across six approaches
in order of development. PPL+CMA-ES with uniform $\alpha$
(orange) and both SAE Projection variants (green) remain below
the $p < 0.05$ threshold ($z = 1.96$, red dashed line) and
$p < 0.01$ threshold ($z = 2.58$, red dotted line). All three
Raw Task Vector configurations (blue) cross into significance,
with our final SP4 14L $\alpha{=}0.80$ reaching $z = +3.41$.
The decisive transition is not feature width or layer count
but whether the task vector is projected through SAE features
before injection.}
\label{fig:method-evolution}
\end{figure}

Figure~\ref{fig:method-evolution} traces the NT $z$-score
across all six approaches in chronological order of development,
from PPL+CMA-ES search through to our final Raw TV SP4
configuration; the transition into significance coincides
exactly with the switch from SAE-projected to raw injection.

\begin{table}[t]
\centering
\caption{Raw task vector vs.\ SAE-projected injection. ``Energy
retained'' denotes
$\|\Delta W_{\text{proj}}\|_F / \|\Delta W\|_F$. All
configurations use the same v2 task vector. The $16\times$
increase in feature width does not close the gap. All $p$\,(NT) values are two-sided, using the same normal
approximation as Table~\ref{tab:main}.}
\label{tab:projection}
\small
\begin{tabular}{lcccc}
\toprule
Method & Energy Retained & NT $z$ & \#Sig & $p$(NT) \\
\midrule
SAE Proj (16K, rank-3)   & $\sim 2.1\%$  & $+1.50$ & 0/7 & $0.1336$ \\
SAE Proj (262K, rank-3)  & $\sim 3.5\%$  & $+0.79$ & 0/7 & $0.4296$ \\
Raw TV (E3, 7 layers)    & $100\%$       & $+2.02$ & 1/7 & $0.0434$ \\
Raw TV (SP4, 14 layers)  & $100\%$       & $+3.41$ & 5/7 & $\mathbf{0.0007}$ \\
\bottomrule
\end{tabular}
\end{table}

SAE projection at the standard 16K feature width retains only a few percent of the modification energy and produces no
statistically significant improvement on any subject (NT
$z = +1.50$, $p = 0.133$). Scaling to $262$K features retains
$\sim 3.5\%$ of the energy but NT performance \emph{decreases}
($z = +0.79$, $p = 0.43$). The bottleneck is the projection
step itself, not feature granularity. This is consistent with
the geometric argument in Section~\ref{sec:method}: SAE decoder
vectors span the activation manifold at layer $l$, a different
geometric object than the column space of $\Delta W^{(l)}$.

\begin{figure}[t]
\centering
\includegraphics[width=1.0\linewidth]{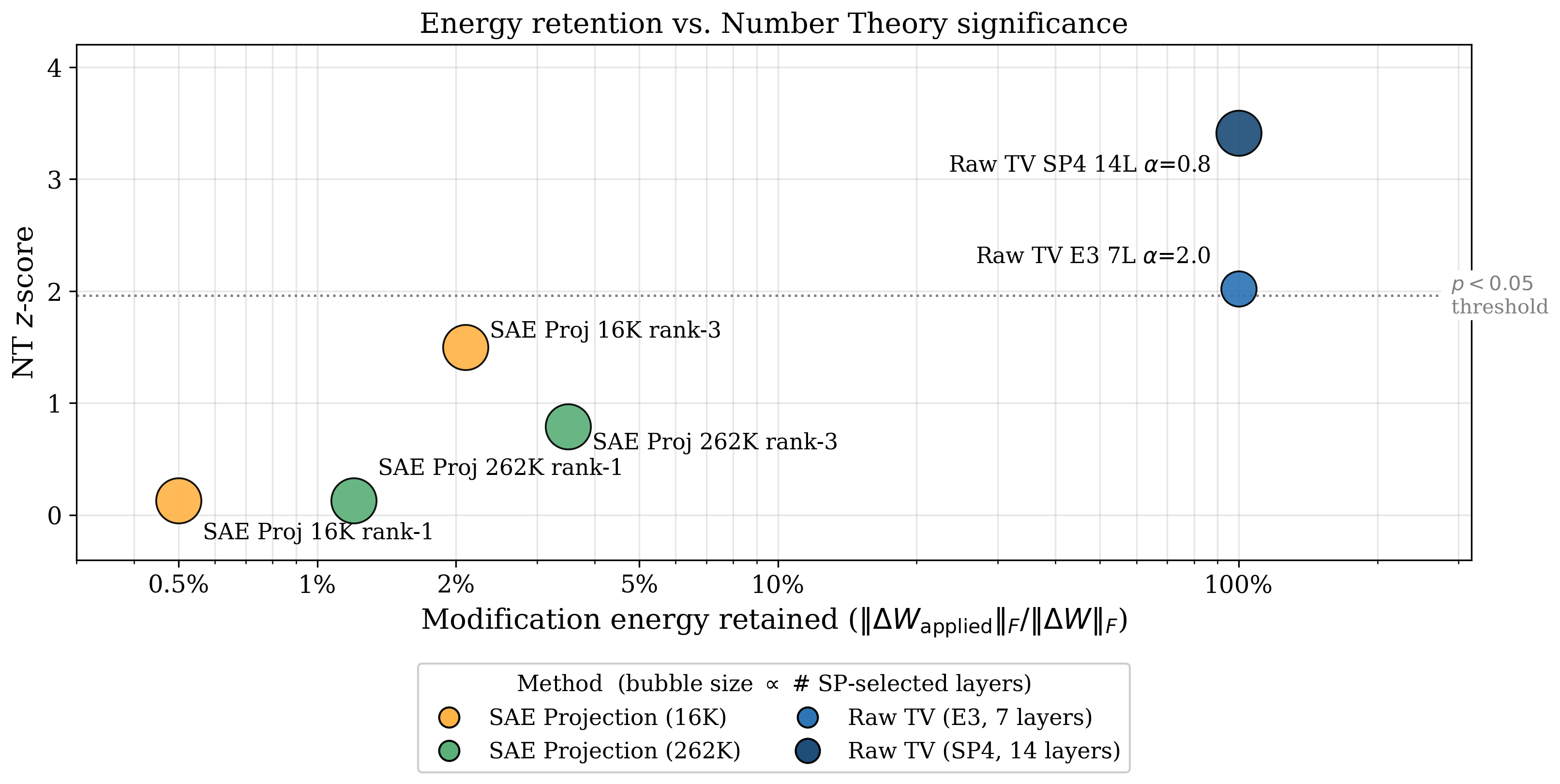}
\caption{Energy retention vs.\ Number Theory significance
($z$-score) across all injection strategies (log-scale $x$-axis).
Orange points: SAE Projection variants (16K and 262K features,
rank-1 and rank-3); teal points: SAE Projection 262K variants;
blue points: Raw Task Vector injection (E3 7-layer and SP4
14-layer). Bubble area is proportional to the number of
SP-selected layers. Both raw-injection configurations
exceed the $p < 0.05$ significance threshold (dotted horizontal
line) while retaining 100\% of $\|\Delta W\|_F$; no
SAE-projected configuration does so at any feature width.
The two-order-of-magnitude gap in energy retention
($\leq 3.5\%$ vs.\ $100\%$) is the proximate cause of
the performance gap.}
\label{fig:energy-vs-z}
\end{figure}

The raw task vector injected into a smaller layer set (7 layers)
achieves $z = +2.02$ ($p = 0.043$)---already outperforming both
SAE-projection configurations despite using half as many layers.
The decisive variable is therefore not how many layers receive
the task vector, but whether the task vector is filtered through
SAE features before injection. Figure~\ref{fig:energy-vs-z} visualises this gap across all
tested configurations: the ${\sim}100\times$ difference in
energy retention between SAE projection ($\leq 3.5\%$) and raw
injection ($100\%$) maps directly onto the difference between
no significant result and $z = +3.41$. Across $30+$ configurations, no SAE-projected variant we tested reached statistical significance
on Number Theory; every raw-injection variant with $\geq 7$
SP-selected layers did. This is strong evidence that SAEs should
be used to identify \emph{where} to edit, but not to filter
\emph{what} to inject.

\section{Analysis and Discussion}
\label{sec:analysis}

\subsection{SAEs as Diagnostic, Not Interventional, Tools}
\label{sec:analysis-diagnosis}

Recent work increasingly suggests that sparse autoencoders are more reliable as probes than as
actuators, and our findings sharpen this pattern in the context of model editing.
The central methodological claim of this work is that SAEs are
useful for \emph{identifying which layers to edit}, but not for
\emph{filtering what to inject into them}. Three lines of
evidence support this: (i) replacing
uniform task-vector application with SAE-derived layer selection moves NT $z$ from $+0.13$ (Full
LoRA merge) to $+3.41$, with 5/7 subjects significantly improved (Section~\ref{sec:results-main}); (ii)
filtering $\Delta W$ through the SAE feature subspace, using the same layer selection, produces no
statistically significant gain on any subject, even at $16\times$ feature width (Section~\ref{sec:results-projection});
and (iii) SAE projection retains only a few percent of $\|\Delta W\|_F$ (up to $\sim 3.5\%$ even at $16\times$ feature width), and the bottleneck is geometric mismatch rather than feature granularity.

This pattern is consistent with a growing body of work on SAE
limitations as intervention tools.
\citet{sharkey2025open} note that SAE reconstruction
systematically loses information, and that SAEs decompose
activations rather than the weights that compute them.
\citet{kantamneni2025sae} show that SAE-based probes do not
consistently outperform simple baselines. \citet{heap2025random}
find that SAEs trained on randomly initialized transformers
produce features qualitatively similar to those from trained
models. \citet{fel2026rabbit} argue that concepts in vision transformers may correspond to convex regions (the "Minkowski Representation Hypothesis") rather than linear directions; if this geometric structure is shared by language models (a hypothesis we cannot directly verify) then linear projection through SAE decoder vectors would be a lossy approximation. Our negative result is consistent with that prediction. Our finding fits this
broader picture: SAEs encode genuine information about
layer-level specialization, but a linear projection onto SAE
features is not a faithful representation of the weight-space
modifications that drive fine-tuned behavior. The practical
implication is a clean separation of roles---SAE for
\emph{where}, raw $\Delta W$ for \emph{what}---which we expect
to generalize beyond mathematical reasoning.

\subsection{Deterministic Evaluation}
\label{sec:analysis-determinism}

We verified determinism by running the SP4 14L
($\alpha = 0.80$) configuration twice with identical
specifications. All seven subject accuracies match at float64
precision (zero difference). Three properties guarantee this:
(i) the task-vector application in Equation~\eqref{eq:inject}
involves only deterministic arithmetic; (ii)
lm-evaluation-harness~\citep{gao2024lmeval} uses greedy decoding
with no sampling; and (iii) the \texttt{math\_verify} metric is
deterministic. The standard errors in our $z$-tests therefore
arise entirely from finite-sample variation in the 540-problem
benchmark, not from evaluation noise. The reported $p = 0.0007$
for Number Theory is a population-level statement, not a
per-run estimate.

\subsection{Additional Observations}
\label{sec:analysis-additional}

Three further observations are documented in detail in the
appendix:
(i) during exploratory CMA-ES~\citep{hansen2006cma} optimization
using perplexity (PPL) as a fitness signal, the lowest-PPL
configuration was not the highest-accuracy one, motivating our
switch to accuracy-based fitness for the rest of the search
(Appendix~\ref{app:ppl});
(ii) combining NT and CP task vectors causes destructive
interference even at modest scaling ($\alpha_{\text{CP}} = 1.0$:
NT $z = -0.74$, ALG $z = -5.61$), suggesting partially
overlapping weight subspaces (Appendix~\ref{app:dual});
and (iii) task-vector quality depends on \emph{domain focus} of
the fine-tuning corpus, not size: a $5\times$-larger
mixed-domain corpus produced essentially no NT effect, while a
focused $4{,}000$-example corpus produced our best result
(Appendix~\ref{app:data}). This mirrors prior editing and merging work that also uses
task accuracy, rather than perplexity, as the primary fitness
signal~\citep{sakana2024evolutionary,ilharco2023editing}.

\section{Limitations}
\label{sec:limitations}

\paragraph{Single base model.}
All experiments use Gemma-3-4B-IT~\citep{gemma3team2025}. Our
mechanistic argument should generalize across architectures, but
we have not verified this empirically. Cross-model validation is
feasible with currently available SAE suites: Llama
Scope~\citep{he2024llamascope} provides full-layer SAE coverage
for Llama-3.1-8B with 32K and 128K feature widths.

\paragraph{Single primary domain.}
The strongest claims concern Number Theory, with positive
transfer to four other math subjects. Independent targeting of
a more feature-dispersed domain (Counting \& Probability,
$\sim 180$ domain-specific features distributed across 18
layers) showed only marginal effects, suggesting the method's
effectiveness depends on whether the target domain has a
sufficiently concentrated SAE-feature signature. Validation on
non-mathematical domains (legal reasoning, code, multilingual
translation) remains untested.

\paragraph{Absolute accuracy ceiling.}
Our edited model achieves $39.4\%$ on Number Theory---a
significant relative gain ($+33\%$) but well below human expert
performance. This reflects the capacity of the 4B base model
rather than a limitation of the editing method.

\paragraph{Task vector prerequisite and SAE availability.}
The method requires one LoRA fine-tuning run
($\sim 4$ GPU-hours, $\sim$\$$16$ at current cloud rates) and
pre-trained SAEs at every layer. Comprehensive public SAE
suites at this scale exist primarily for the Gemma family
(Gemma Scope~\citep{lieberum2024gemma} and Gemma Scope
2~\citep{deepmind2025gemmascope2}) and the Llama
family~\citep{he2024llamascope}. Training new SAEs is
expensive (Gemma Scope used $\sim 15\%$ of Gemma 2 9B's training
budget~\citep{lieberum2024gemma}), but this cost is amortized:
the same SAEs support unlimited downstream layer-selection
queries. As with any capability-enhancing technique, our method could in
principle be applied to harmful domains; we do not explore such
applications and restrict ourselves to mathematical reasoning.

\paragraph{Statistical power.}
Subject-level $z$-tests are well-powered to detect $\sim 4$ pp
effects on every subject given the full Minerva Math test set sizes
($n = 540$ NT; $n \in [474, 1{,}187]$ others). Intermediate
Algebra ($z = +1.65$, $p = 0.099$, two-sided) does not reach
significance and should be interpreted accordingly. PreCalculus
($z = 0.00$) shows no effect under this task vector.

\section{Conclusion}
\label{sec:conclusion}

We presented an interpretability-guided method for selective
model editing that uses Sparse Autoencoder feature analysis to
identify domain-specialized layers, then injects raw
LoRA-derived task vectors into only those layers. On
Gemma-3-4B-IT targeting Number Theory, the method achieves
$z = +3.41$ ($p = 0.0007$), improving 5 of 7 math subjects with
no degradation, and is fully deterministic at float64 precision.

The central scientific finding is methodological: \textbf{SAEs
are useful as diagnostic tools for layer selection, not as
intervention tools for filtering weight-space modifications.}
Projecting task vectors through SAE feature subspaces---the
intuitively obvious approach---retains only a few percent ($\leq 3.5\%$) of the
modification energy and produces no statistically significant
gain on any subject, even at $16\times$ feature width. The
geometric reason is that SAE decoder vectors span the
activation manifold while task vectors live in weight space;
the two geometries are not interchangeable. The near-total energy loss is consistent with this structural incompatibility/geometric mismatch. Further, this finding is
consistent with a growing body of work documenting SAE
limitations as intervention
tools~\citep{sharkey2025open,kantamneni2025sae,heap2025random}
and contributes a sharp empirical demonstration of where the
activation-space/weight-space distinction matters. Our results suggest that the efficacy of this diagnostic selection is contingent on the "sharpness" of the task vector itself: 
a $5\times$ larger mixed-domain corpus (v3) failed to produce a significant effect, reinforcing that surgical editing requires both precise layer selection and high-fidelity, domain-focused training data.

Future work includes cross-model validation using Llama
Scope~\citep{he2024llamascope}, extension to non-mathematical
domains, integration with interference-mitigation strategies for
multi-domain editing~\citep{yadav2023ties,he2025localize}, and a
more principled characterization of when activation-space SAE
analysis transfers cleanly to weight-space interventions versus
when it fails. The broader implication is that interpretability
tools and editing tools may serve complementary rather than
substitutable roles---a separation we expect to generalize
beyond the specific setting studied here.

\newpage
\bibliographystyle{unsrtnat}    
\bibliography{references}        


\clearpage
\appendix

\section{Reproduction Commands}
\label{app:reproduction}

\begin{verbatim}
# Step 1: LoRA fine-tuning (produces task vector v2)
python experiments/nt_train_lora_v2.py \
    --gpu 0 --name lora_v2 --epochs 5 --lora_r 16

# Step 2: Compute task vector and apply to SP>=4.0 layers
python experiments/raw_tv_sweep_v2.py \
    --config SP4_14L --alpha 0.80 --tv v2

# Step 3: Evaluate with lm-eval
lm_eval --model hf \
    --model_args pretrained=results/raw_tv_v2/models/SP4_14L_a0.8 \
    --tasks minerva_math_* --num_fewshot 4 --batch_size 4
\end{verbatim}

The commands above document the script-level invocation pattern
used in our experiments. Implementation code is proprietary and
is not released with this submission. The methodology is fully
specified by Section~\ref{sec:setup} (training hyperparameters),
Section~\ref{sec:method} (algorithm), and
Appendix~\ref{app:sp-scores} (per-layer specificity scores),
enabling independent re-implementation against publicly
available assets.

\section{Alpha Response Curve: Full Sweep}
\label{app:alpha}

\begin{figure}[h]
\centering
\includegraphics[width=1.0\linewidth]{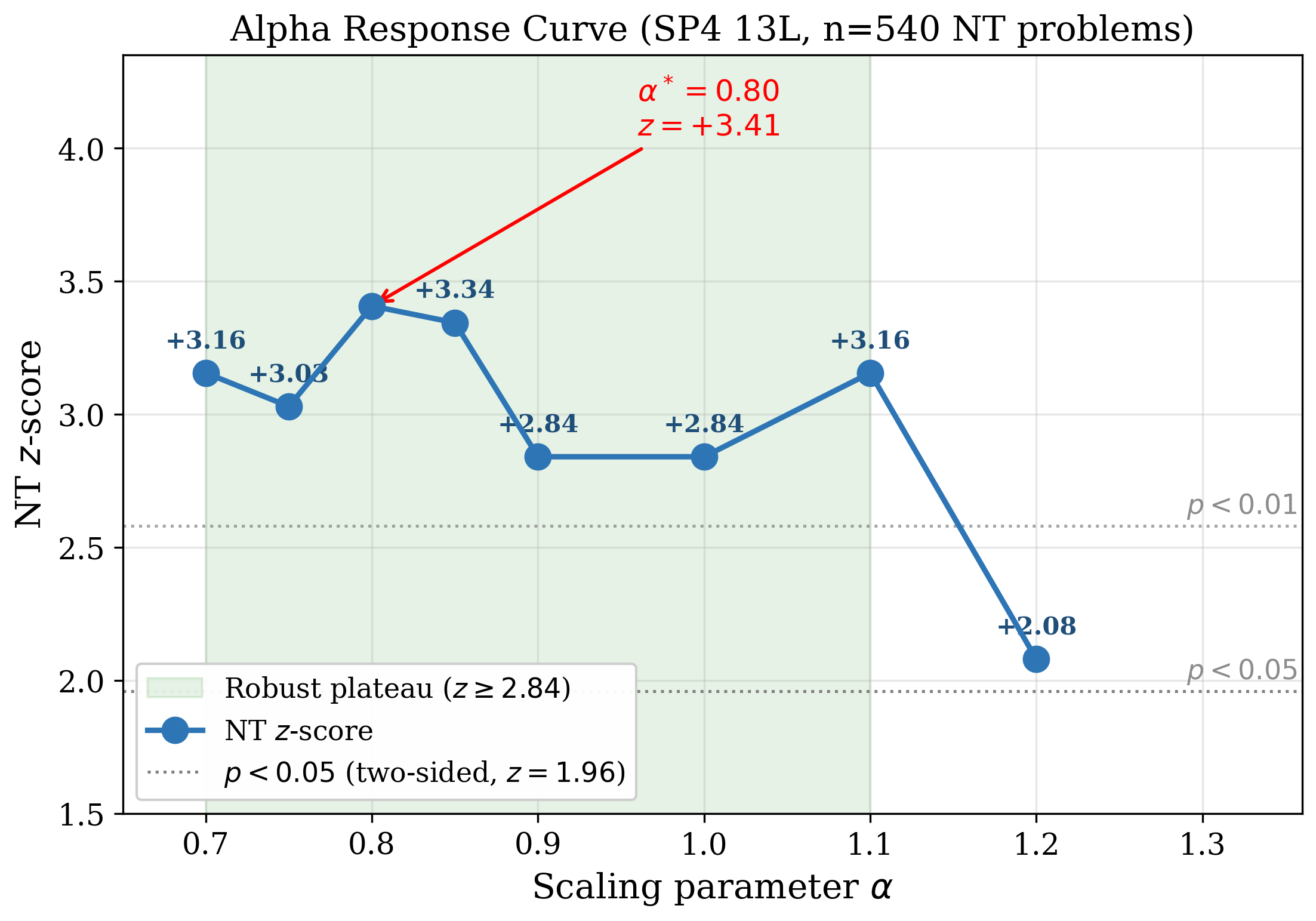}
\caption{Number Theory $z$-score as a function of $\alpha$ for
the SP4~14L configuration ($n = 540$ NT problems). The green
shaded region marks the robust plateau ($z \geq 2.84$,
$\alpha \in [0.70, 1.10]$). The optimal $\alpha^* = 0.80$
($z = +3.41$) is annotated in red. Two dotted reference lines
indicate the $p < 0.05$ ($z = 1.96$) and $p < 0.005$
($z = 2.58$) thresholds. Performance drops sharply only at
$\alpha = 1.20$, marking the onset of over-modification.}
\label{fig:alpha-curve}
\end{figure}

\begin{figure}[h]
\centering
\includegraphics[width=1.0\linewidth]{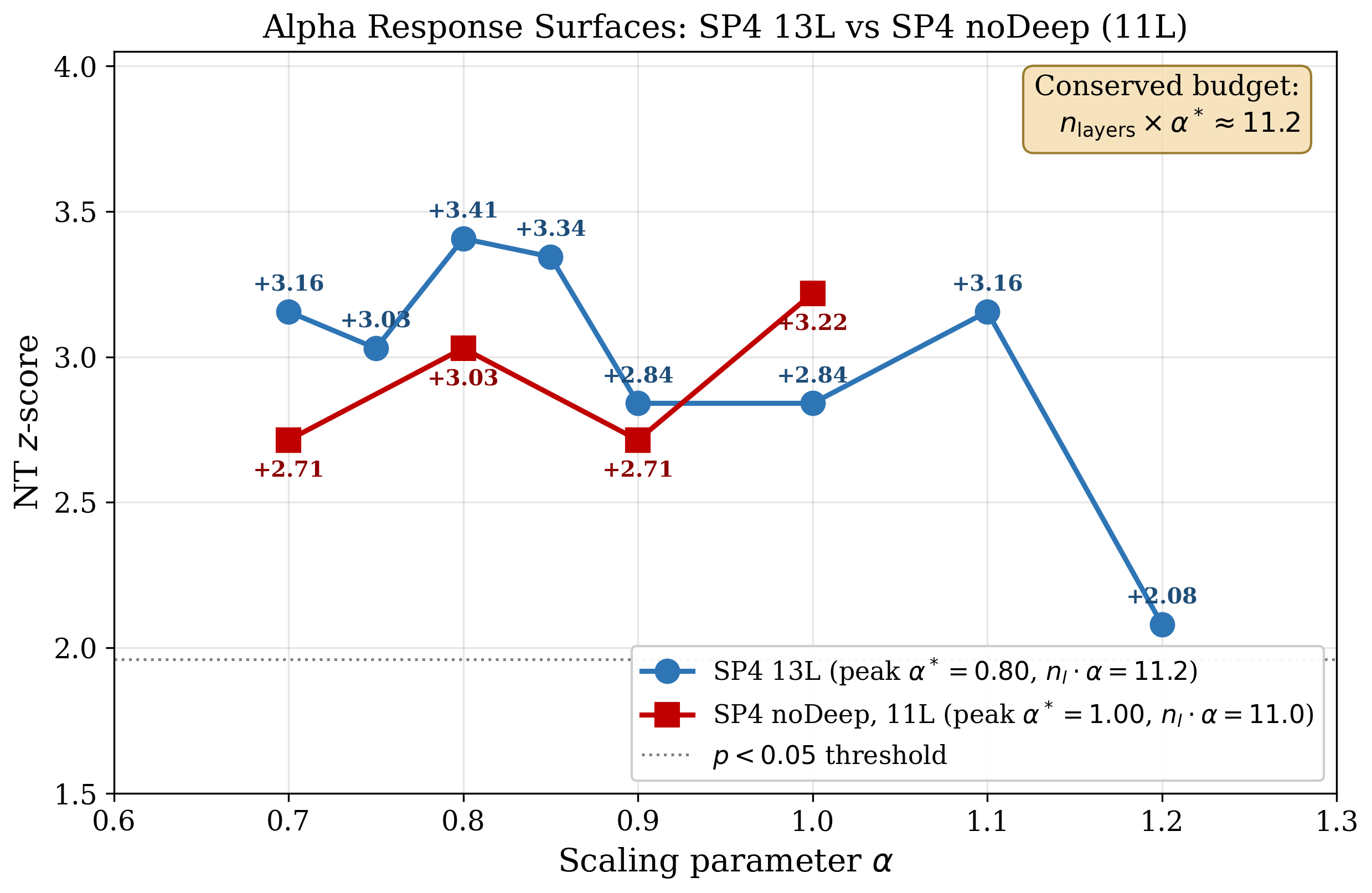}
\caption{Alpha response surfaces for SP4~14L (blue circles,
14 layers) and SP4~noDeep (red squares, 11 layers). Both
configurations peak within the same $z$-score range despite
different layer counts: SP4~14L at $\alpha^* = 0.80$
($n_l \times \alpha^* = 11.2$) and SP4~noDeep at
$\alpha^* = 1.00$ ($n_l \times \alpha^* = 11.0$). The near identical products
($n_{\text{layers}} \times \alpha^* \approx 11.2$) are consistent with the interpretation that
$\alpha$ compensates for reduced layer coverage rather than
independently controlling edit strength.}
\label{fig:alpha-two-configs}
\end{figure}

Table~\ref{tab:alpha-full} reports the full alpha sweep for the
SP4~14L configuration referenced in
Section~\ref{sec:results-alpha}.
Figures~\ref{fig:alpha-curve} and~\ref{fig:alpha-two-configs}
visualize the response surface for SP4~14L alone and in
comparison with SP4~noDeep (11 layers).

\begin{table}[h]
\centering
\caption{Alpha response curve for the SP4 14L configuration. NT
accuracy and $z$-scores as a function of $\alpha$. The response
surface is robust over a wide range ($z \geq 2.84$ for
$\alpha \in [0.70, 1.10]$); only $\alpha = 1.20$ shows
substantial degradation.}
\label{tab:alpha-full}
\small
\begin{tabular}{lcccccc}
\toprule
$\alpha$ & NT Acc & NT $z$ & CP $z$ & ALG $z$ & GEO $z$ & \#Sig \\
\midrule
0.70 & 38.7\% & $+3.16$ & $+2.5$ & $+3.0$ & $+2.5$ & 4/7 \\
0.75 & 38.3\% & $+3.03$ & $+2.5$ & $+2.8$ & $+2.0$ & 5/7 \\
0.80 & 39.4\% & $+3.41$ & $+2.6$ & $+2.9$ & $+2.2$ & 5/7 \\
0.85 & 39.3\% & $+3.34$ & $+2.7$ & $+2.9$ & $+2.6$ & 5/7 \\
0.90 & 37.8\% & $+2.84$ & $+2.8$ & $+3.2$ & $+2.7$ & 5/7 \\
1.10 & 38.7\% & $+3.16$ & $+2.7$ & $+2.8$ & $+1.8$ & 4/7 \\
1.20 & 35.6\% & $+2.08$ & $+2.6$ & $+2.4$ & $+2.3$ & 5/7 \\
\bottomrule
\end{tabular}
\end{table}

\clearpage
\section{Layer Selection: Per-Subject Heatmap}
\label{app:layers}

\begin{figure}[h]
\centering
\includegraphics[width=1.0\linewidth]{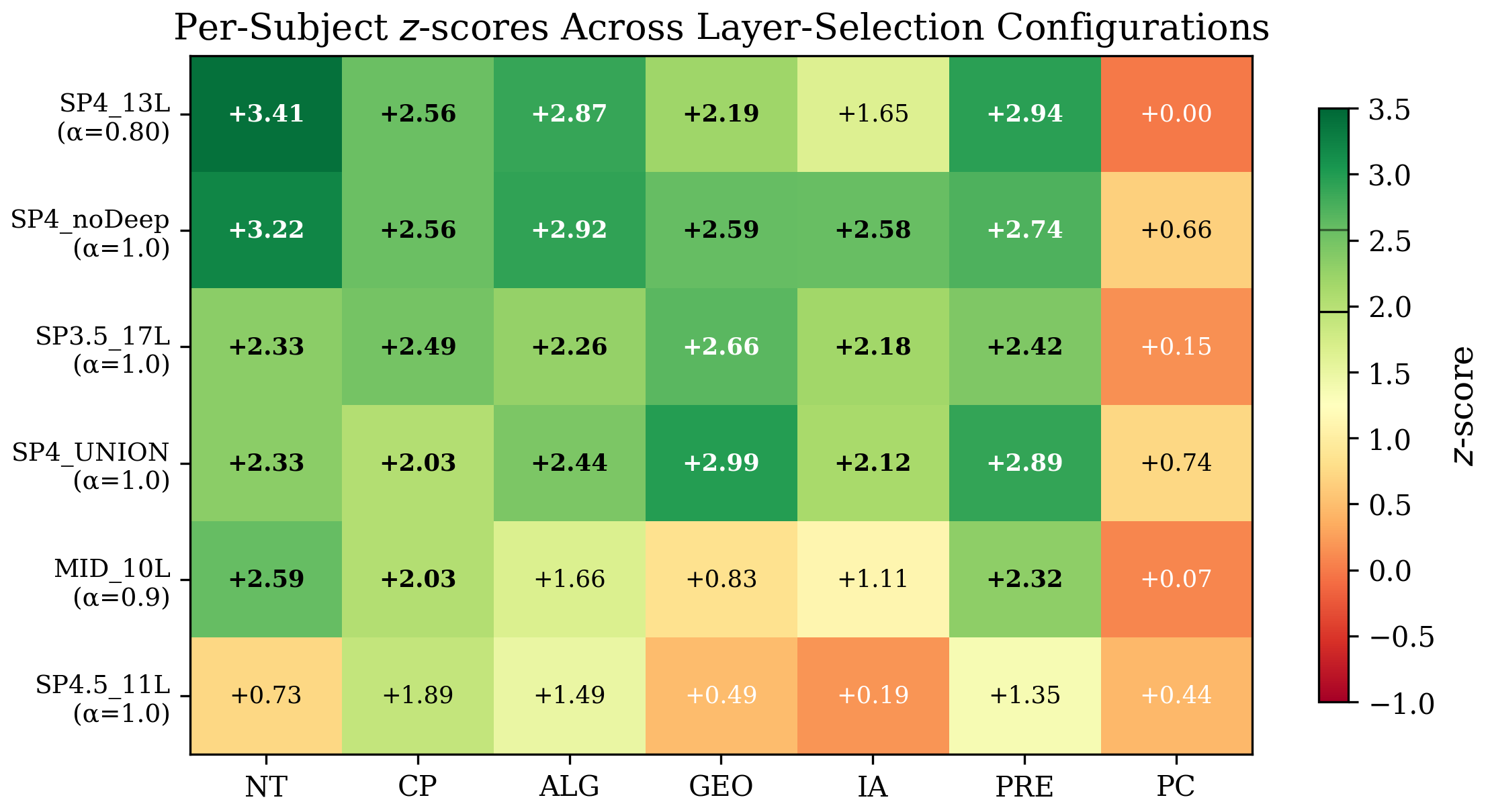}
\caption{Per-subject $z$-scores across six layer-selection
configurations. SP4 14L and SP4 noDeep dominate across most
subjects; SP4.5 11L collapses to non-significance despite a
similar layer count, showing that the SP threshold acts as a
categorical separator rather than a smooth knob.}
\label{fig:layer-heatmap}
\end{figure}

Figure~\ref{fig:layer-heatmap} shows per-subject $z$-scores for
the six representative layer-selection configurations
referenced in Section~\ref{sec:results-layers}.

\section{Perplexity as a Fitness Metric}
\label{app:ppl}

\begin{figure}[h]
\centering
\includegraphics[width=1.0\linewidth]{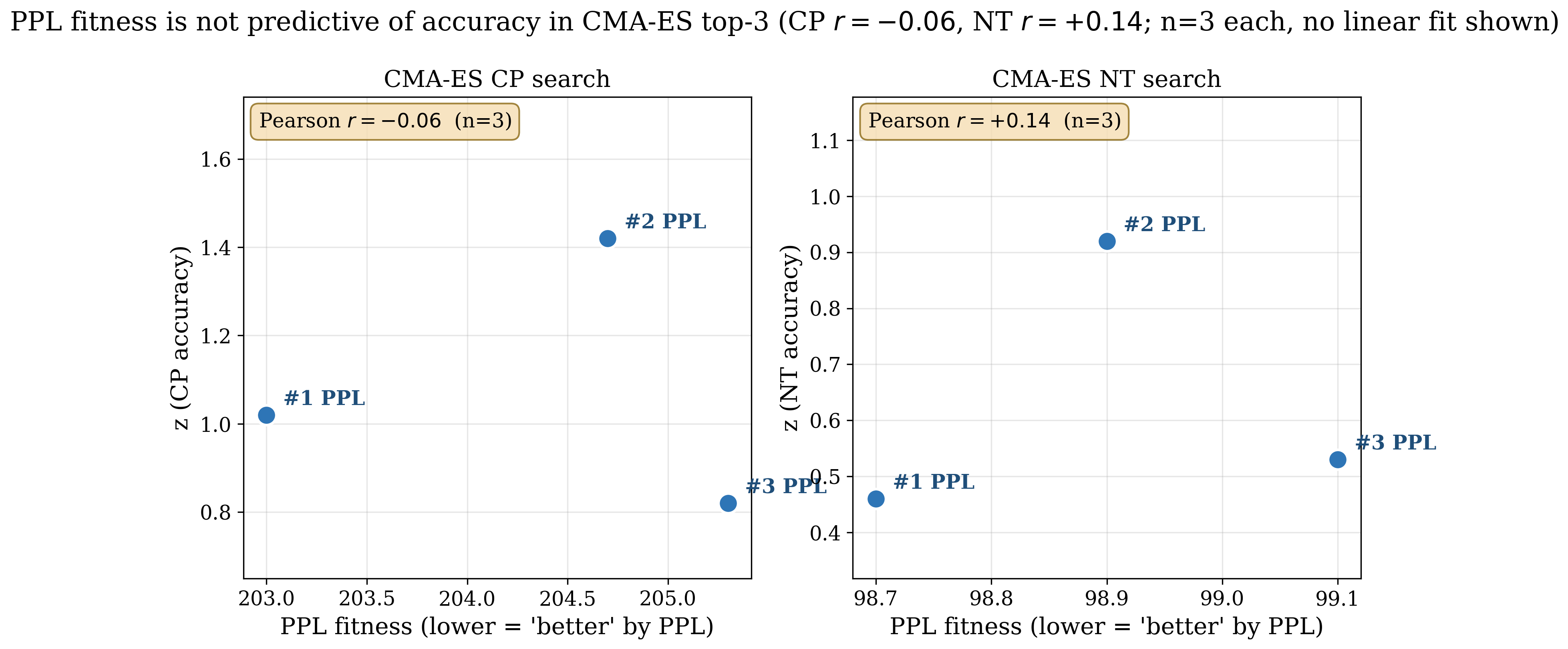}
\caption{PPL fitness vs.\ accuracy $z$-score for the top-3
CMA-ES configurations in the CP search (left) and NT search
(right). Each point is one configuration ranked by PPL
fitness (lower PPL = ``better'' by the fitness signal).
Red dashed lines show the Pearson regression fit.
CP: $r = -0.06$ (essentially uncorrelated); NT: $r = +0.14$
(slightly positive, opposite to the anti-correlation the
fitness signal assumes). In both searches the
lowest-PPL configuration (\#1 PPL) is \emph{not} the
highest-accuracy one, motivating our switch to
accuracy-based fitness for the remainder of the search.}
\label{fig:ppl-scatter}
\end{figure}

During an exploratory CMA-ES~\citep{hansen2006cma} optimization
of per-layer $\alpha$ values ($\sim 60$ GPU-hours), we used
held-out perplexity (PPL) as a proxy fitness signal and
subsequently evaluated the top configurations on accuracy. The
top three configurations spanned only $\sim 0.4$ PPL points but
differed by $\sim 1.3$ percentage points in NT accuracy, and
the lowest-PPL configuration was not the highest-accuracy one.
With $n = 3$ configurations in this range, the within-search
correlation is too weak to support a generalizable claim, but
the observation was sufficient to motivate switching to
accuracy-based fitness. Figure~\ref{fig:ppl-scatter} shows the
PPL--accuracy relationship for both the CP and NT CMA-ES
searches: Pearson $r = -0.06$ for CP and $r = +0.14$ for NT,
both consistent with no reliable correlation.

The structural reason PPL and accuracy can diverge is that PPL
is a token-level average over a corpus distribution, whereas
mathematical accuracy is a sequence-level boolean over
chain-of-thought completions. A model can produce fluent but
incorrect reasoning chains and achieve low PPL; conversely, a
model that solves problems correctly may produce slightly more
``surprising'' tokens than the base distribution. Our practical
conclusion, which is to use accuracy directly as the fitness
metric, is consistent with prior practice in evolutionary
merging~\citep{sakana2024evolutionary} and task
arithmetic~\citep{ilharco2023editing}. Whether PPL is a
reliable fitness signal for non-mathematical model editing
remains an open question.

\section{Cross-Domain Interference in Multi-Task Injection}
\label{app:dual}

We tested whether two domain-specific task vectors could be
combined additively. Combining the NT task vector with a
Counting \& Probability (CP) task vector at modest scaling
($\alpha_{\text{CP}} = 0.5$) reduced NT performance from
$z = +3.41$ to $z = +2.65$ (3/7 significant), while increasing
CP scaling to $\alpha_{\text{CP}} = 1.0$ caused catastrophic
collapse: NT $z = -0.74$ (accuracy roughly unchanged). Under DUAL (SP4 NT + CP task vectors, both at $\alpha{=}1.0$), Algebra
(\texttt{math\_verify}) falls from $\mathbf{61.3\%}$ to $\mathbf{\sim 50.0\%}$ (base model vs.\ edited model on the same 1{,}187-problem ALG split), with $z_{\text{ALG}}{=}{-}5.61$. This indicates that NT and CP task vectors occupy partially overlapping weight subspaces and cannot be simply summed without destructive
interference.

This is consistent with prior findings on task-vector
interference~\citep{yadav2023ties,he2025localize} and motivates
integration with existing interference-mitigation strategies.
The SP-based layer selection identifies $\geq 14$ layers per
domain, and these sets overlap substantially across NT and CP.
Future work could combine SAE-guided layer selection with
TIES-style sign-conflict resolution~\citep{yadav2023ties} or
DARE-style random dropping~\citep{yu2024language}.

\section{Task Vector Source: Data Quality Over Quantity}
\label{app:data}

We compared task vectors trained on three datasets:

\begin{itemize}[topsep=2pt,itemsep=2pt,leftmargin=*]
  \item \textbf{v2}: $3{,}865$ NT-focused samples $\rightarrow$
        NT $z = +3.41$ (5/7 significant).
  \item \textbf{v25}: $6{,}155$ samples (NT-focused, broader)
        $\rightarrow$ NT $z = +2.84$, ALG $z = +4.07$.
  \item \textbf{v3}: $20{,}000$ mixed-domain samples $\rightarrow$
        NT $z = +0.46$ (0/7 significant).
\end{itemize}

More data does not produce better task vectors for the target
subject. The $5\times$-larger mixed-domain corpus (v3) yields a
task vector essentially indistinguishable from no edit on
Number Theory. The intermediate v25 corpus ($1.6\times$ the v2
size, broader content) trades NT performance for stronger
transfer to Algebra. This indicates that the domain focus of
the fine-tuning corpus, not its size, determines the
target-domain content of the resulting task vector.

\section{Configuration Rankings}
\label{app:configs}

Table~\ref{tab:full-configs} reports the top 10 configurations
across all $30+$ ablations evaluated with full lm-eval (540
problems, math\_verify metric). All configurations use the v2
task vector unless noted.

\begin{table}[h]
\centering
\caption{Top-10 configurations by NT $z$-score. ``DUAL''
denotes combined NT and CP task vectors. ``v25'' denotes the
6155-sample task vector variant.}
\label{tab:full-configs}
\small
\begin{tabular}{rlcccl}
\toprule
Rank & Configuration & $\alpha$ & NT $z$ & \#Sig & Note \\
\midrule
1  & SP4 14L            & 0.80 & $+3.41$ & 5/7 & Global optimum \\
2  & SP4 noDeep         & 1.00 & $+3.22$ & 6/7 & Multi-subject best \\
3  & SP4 14L            & 0.70 & $+3.16$ & 4/7 & \\
3  & SP4 14L            & 1.10 & $+3.16$ & 4/7 & \\
5  & DUAL MID+CP0.5     & 1.00 & $+3.09$ & 4/7 & DUAL config \\
6  & SP4 14L            & 0.90 & $+2.84$ & 5/7 & \\
6  & SP4 14L            & 1.00 & $+2.84$ & 5/7 & \\
6  & v25 SP4 14L        & 1.00 & $+2.84$ & 5/7 & 6155-sample TV \\
9  & MID 10L            & 0.90 & $+2.59$ & 3/7 & \\
9  & MID 10L            & 1.00 & $+2.59$ & 2/7 & \\
\bottomrule
\end{tabular}
\end{table}

\section{Per-Layer Specificity Scores}
\label{app:sp-scores}

Table~\ref{tab:sp_scores} lists the number of domain-specific
features (specificity $> 1.0$) per layer for Number Theory, as
identified by Gemma Scope 2~\citep{deepmind2025gemmascope2} with 16K
features per layer. Layers with $\mathrm{SP} \geq 4.0$
(selected for injection) are marked.

\begin{table}[h]
\centering
\caption{Per-layer Number Theory specificity score
$\mathrm{SP}(l) = \max_j \mathrm{spec}(l, j)$
computed using Gemma Scope 2 SAEs with 16K features per layer.
Layers with $\mathrm{SP}(l) \geq 4.0$ (selected for raw task vector injection)
are marked with $\bullet$.
The companion column "\#feat" shows the number of features with $\mathrm{spec}(l,j) > 1.0$
in each layer (used in Figure~\ref{fig:sp_scores} bottom panel).}
\label{tab:sp_scores}
\begin{tabular}{cccc|cccc}
\toprule
Layer & SP & \#feat & Sel. & Layer & SP & \#feat & Sel. \\
\midrule
L6  & 1.21 & 1  &           & L20 & 7.00 & 9  & $\bullet$ \\
L7  & 1.10 & 1  &           & L21 & 4.75 & 12 & $\bullet$ \\
L9  & 1.08 & 2  &           & L22 & 6.30 & 11 & $\bullet$ \\
L10 & 2.30 & 4  &           & L23 & 5.58 & 6  & $\bullet$ \\
L11 & 1.23 & 7  &           & L24 & 4.21 & 8  & $\bullet$ \\
L12 & 1.90 & 6  &           & L25 & 5.35 & 8  & $\bullet$ \\
L13 & 3.47 & 5  &           & L26 & 3.83 & 8  &           \\
L14 & 4.07 & 20 & $\bullet$ & L27 & 4.88 & 4  & $\bullet$ \\
L15 & 4.09 & 24 & $\bullet$ & L28 & 3.52 & 12 &           \\
L16 & 3.74 & 21 &           & L29 & 3.37 & 12 &           \\
L17 & 5.08 & 22 & $\bullet$ & L30 & 5.54 & 13 & $\bullet$ \\
L18 & 2.33 & 18 &           & L31 & 8.80 & 13 & $\bullet$ \\
L19 & 7.82 & 16 & $\bullet$ & L32 & 5.02 & 12 & $\bullet$ \\
\bottomrule
\end{tabular}
\end{table}

\section{Broader Impacts}
\label{app:impacts}

This work develops methods for surgical model editing of
existing language models. We see two principal positive
implications: (i) the method enables targeted capability
enhancement at substantially lower compute cost than full
fine-tuning, reducing the carbon footprint of model
specialization; (ii) the negative result on SAE projection
clarifies what interpretability tools can and cannot reliably
do for editing applications, supporting more rigorous use of
mechanistic-interpretability tooling in practice.

We see one principal risk worth flagging: targeted model
editing techniques can in principle be applied to enhance
capabilities for harmful purposes, in the same way that any
fine-tuning method can. Our method is no more dual-use than
LoRA fine-tuning itself, which is widely available. We do not
release new model weights, datasets, or capabilities beyond
what is already publicly accessible.

\section{Licenses for Used Assets}
\label{app:licenses}

This work uses the following publicly available assets:
\begin{itemize}[topsep=2pt,itemsep=2pt,leftmargin=*]
\item Gemma-3-4B-IT model~\citep{gemma3team2025}: Gemma Terms
      of Use (\url{https://ai.google.dev/gemma/terms}).
\item Gemma Scope 2 SAEs~\citep{deepmind2025gemmascope2}: CC-BY-4.0
      (per the Hugging Face release).
\item MATH dataset~\citep{hendrycks2021math}: MIT License.
\item lm-evaluation-harness~\citep{gao2024lmeval}: MIT License.
\end{itemize}
We use each asset within the scope permitted by its respective
license.


\end{document}